\documentclass[sn-mathphys-num]{sn-jnl}

\usepackage{graphicx}%
\usepackage{multirow}%
\usepackage{amsmath,amssymb,amsfonts}%
\usepackage{amsthm}%
\usepackage{mathrsfs}%
\usepackage[title]{appendix}%
\usepackage{xcolor}%
\usepackage{textcomp}%
\usepackage{manyfoot}%
\usepackage{booktabs}%
\usepackage{algorithm}%
\usepackage{algorithmicx}%
\usepackage{algpseudocode}%
\usepackage{listings}%

\usepackage{bm}
\usepackage{tabularx}
\usepackage{multirow}
\usepackage{xspace}
\usepackage{hyperref}

\newcommand{\subf}[2]{%
  {\footnotesize\begin{tabular}[t]{@{}c@{}}
  #1\\#2
  \end{tabular}}%
}

\newcommand{\ours}{$\texttt{DiffER}$\xspace}

\begin{document}

\title[{\ours}: Categorical Diffusion Models for Chemical Retrosynthesis]{{\ours}: Categorical Diffusion Models for Chemical Retrosynthesis}


\author*[1]{\fnm{Sean} \sur{Current}}\email{current.33@osu.edu}

\author[1]{\fnm{Ziqi} \sur{Chen}}\email{chen.8484@osu.edu}

\author[2]{\fnm{Daniel} \sur{Adu-Ampratwum}}\email{adu-ampratwum.1@osu.edu}

\author[1,2,3,4]{\fnm{Xia} \sur{Ning}}\email{ning.104@osu.edu}

\author*[1]{\fnm{Srinivasan} \sur{Parthasarathy}}\email{srini@cse.ohio-state.edu}

\affil[1]{\orgdiv{Computer Science and Engineering}, \orgname{The Ohio State University}, \orgaddress{\city{Columbus}, \postcode{43210}, \state{OH}, \country{USA}}}

\affil[2]{\orgdiv{College of Pharmacy}, \orgname{The Ohio State University}, \orgaddress{\city{Columbus}, \postcode{43210}, \state{OH}, \country{USA}}}

\affil[3]{\orgdiv{Translational Data Analytics Institute}, \orgname{The Ohio State University}, \orgaddress{\city{Columbus}, \postcode{43210}, \state{OH}, \country{USA}}}

\affil[4]{\orgdiv{Biomedical Informatics}, \orgname{The Ohio State University}, \orgaddress{\city{Columbus}, \postcode{43210}, \state{OH}, \country{USA}}}


\abstract{Methods for automatic chemical retrosynthesis have found recent success through the application of models traditionally built for natural language processing, primarily through transformer neural networks. These models have demonstrated significant ability to translate between the SMILES encodings of chemical products and reactants, but are constrained as a result of their autoregressive nature. We propose \ours, an alternative template-free method for retrosynthesis prediction in the form of categorical diffusion, which allows the entire output SMILES sequence to be predicted in unison. We construct an ensemble of diffusion models which achieves state-of-the-art performance for top-1 accuracy and competitive performance for top-3, top-5, and top-10 accuracy among template-free methods. We prove that \ours is a strong baseline for a new class of template-free model and is capable of learning a variety of synthetic techniques used in laboratory settings. \textbf{Scientific Contribution.} \ours represents a novel template-free technique for single-step retrosynthesis prediction which is able to outperform a variety of other template-free methods on top-k accuracy metrics. By constructing an ensemble of categorical diffusion models with a novel length prediction component with variance, our method is able to approximately sample from the posterior distribution of reactants, producing results with strong metrics of confidence and likelihood. Furthermore, our analyses demonstrate that accurate prediction of the SMILES sequence length is key to further boosting the performance of categorical diffusion models.}




\maketitle


Retrosynthesis prediction is a vital step in organic synthesis tasks, particularly those posed for drug discovery or drug engineering. In forward synthesis prediction, the products of a chemical reaction are predicted from known reactants; retrosynthesis prediction reverses this process, instead predicting possible reactants that would produce a target product. Repeated application of retrosynthesis prediction can help chemists construct synthetic pathways for drug targets \cite{segler2017neural, segler2018planning, shen2021automation}, promoting the discovery and advancement of new pharmaceuticals. While numerous types of models for computer-aided retrosynthesis have been proposed, many modern approaches utilize data-driven machine learning to find suitable models for retrosynthesis prediction.

Recent advances in machine learning models for chemical retrosynthesis have taken advantage of transformer architectures originally crafted for natural language tasks. Instead of operating on natural language, these models are set to operate on SMILES \cite{weininger1988smiles} encodings of chemical products and reactants, effectively translating between two molecular sets \cite{liu2017retrosynthetic}. Indeed, recent work has done precisely this, going as far as using neural machine translation models originally built for language tasks for retrosynthesis prediction \citep{zhong2022root, irwin2022chemformer}. A more detailed overview of current work in single-step retrosynthesis is provided in \ref{sec:related}. While these models exhibit remarkable performance compared to other baseline methods for retrosynthesis, they impose autoregressive constraints during training and prediction, enforcing sequential decoding of the SMILES string.

\begin{figure}
    \small
    \centering
    \includegraphics[width=\linewidth]{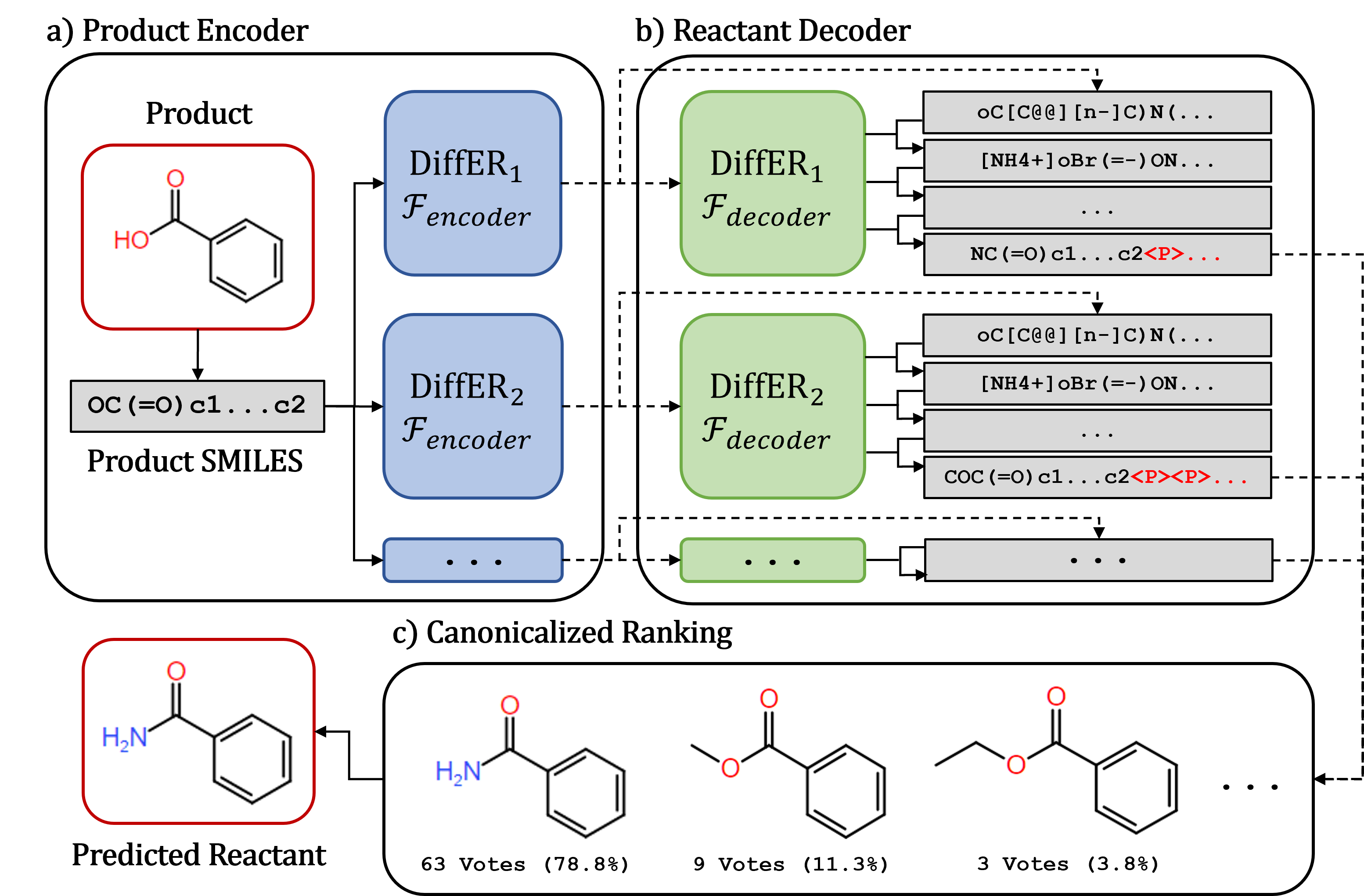}
    \caption{Schematic of \ours, our diffusion process for retrosynthesis. \textbf{a)} The Product Encoder, which converts the product to a random, root-aligned SMILES form and encodes it using the encoder portion of the \ours models. The encoder models also predict the size of the diffusion noise vector. \textbf{b)} The Reactant Decoder, which diffuses the reactant from randomly initialized categorical noise using the decoder portion of the \ours models. Decoder models can predict pad tokens \texttt{<P>} to diffuse SMILES strings of varying lengths. \textbf{c)} The Canonicalized Ranking, which ranks the canonical molecules predicted by the \ours models by the number of times they are generated. The top-ranked reactant is the predicted reactant.}
    \label{fig:schema}
\end{figure}

In this work, we offer \ours, an alternative to sequential decoding of SMILES strings in the prediction process by utilizing categorical diffusion models rather than autoregressive models. This exchange allows \ours to decode the entire SMILES string in unison, rather than in an autoregresssive manner. We hypothesize that by allowing the model to predict the entire SMILES string in unison, it may better learn structural relationships within the molecule. Additionally, diffusion models have experienced a recent surge of success for many generation tasks across domains. By employing categorical diffusion for the chemical retrosynthesis task, we provide a strong and competitive baseline for a new class of model. An outline of our approach is display in Figure \ref{fig:schema}, while a more detailed overview is available in Section \ref{sec:methods}. \ours achieves state-of-the-art top-1 accuracy and competitive performance for top-3, top-5, and top-10 accuracy among template-free methods.

Our main contributions to the domain of computer-aided chemical retrosynthesis are as follows:
\begin{itemize}
    \item We develop and test \ours, an ensemble of diffusion models for the chemical retrosynthesis task and compare our performance against a variety of existing methods, achieving state-of-the-art top-1 accuracy and competitive top-3 and top-5 accuracy.
    \item We implement a novel method for estimating the length of the output sequence prior to prediction so that the diffusion process can be properly initialized, while accounting for possible error and variance in the predicted sequence length.
    \item We compare against a diffusion model with oracle length prediction to demonstrate an experimental upper-limit of the diffusion approach if the output sequence length is able to be perfectly estimated. We find that the oracle model greatly outperforms all existing methods for computer-aided retrosynthesis, demonstrating the viability of diffusion models for this task given a target sequence length.
    \item We also include results for a baseline diffusion model with length prediction which does not account for variance or error in the output length, and show that our training and sampling methodology greatly improves the performance of diffusion models.
    \item We demonstrate the results of our method on a set of four reactions, highlighting the benefits of our approach as well as the limitations for each reaction.
    \item We discuss the current advantages and limitations of our diffusion approach and identify directions of future research to improve upon our work.
\end{itemize}

\section{Methods}
\label{sec:methods}



We define the retrosynthesis task as a sequence-to-sequence modeling problem and combine the approaches of Hoogeboom et al. \cite{hoogeboom2021argmax}, DiffuSeq \cite{gong2022diffuseq, gong2023diffuseq}, and MaskPredict \cite{ghazvininejad2019mask} to train conditional, multinomial diffusion models for the retrosynthesis task. We utilize a traditional encoder-decoder transformer architecture to construct individual conditional diffusion models, where the encoder is responsible for learning the representation of the conditional features and the decoder acts as the diffusion approximation model. In the end, eight individual diffusion models are ensembled to produce \ours, a novel ensemble model for chemical retrosynthesis.

Let $\bm{x}_t \in \{0,1\}^{K \times \ell_x}$ be a sequence of $\ell_x$ one-hot encoded vectors of size $K$ representing the source and let $\bm{y}_t \in \{0,1\}^{K \times \ell_y}$ be similarly defined representing the target. Let $\mathcal{C}$ denote a categorical probability distribution with parametrized probability. Finally, let $\mathcal{F}$ represent a transformer neural network with encoder $\mathcal{F}_{\text{encoder}}$, decoder $\mathcal{F}_{\text{decoder}}$, and sequence length prediction classifier $\mathcal{F}_{\text{length}}$.

\subsection{Forward Noising Process}
Following Hoogeboom et al. \cite{hoogeboom2021argmax}, we define the multinomial diffusion process $q$ as a categorical distribution $\mathcal{C}$ with probability $\beta_t$ of sampling uniformly from $\{0, 1\}^K$ for any step in the sequence, and apply noising to the target matrix $y_t$:
\begin{equation}
    q(\bm{y}_t | \bm{y}_{t-1}) = \mathcal{C}(\bm{y}_t | (1-\beta_t)\bm{y}_{t-1} + \beta_t / K).
\end{equation}
This forms a Markov chain allowing $\bm{y}_t$ to be directly sampled from $\bm{y}_0$:
\begin{equation}
    q(\bm{y}_t | \bm{y}_0) = \mathcal{C}(\bm{y}_t | \bar{\alpha}_t\bm{y}_0 + (1 - \bar{\alpha}_t)/K),
\end{equation}
where $\alpha_t = 1 - \beta_t$ and $\bar{\alpha}_t = \prod_{\tau=1}^t\alpha_\tau$. Note that in the forward noising process, we only apply noise to the target sequence $\bm{y}_0$.

\subsection{Conditional Denoising}
We continue following the work of Hoogeboom et al. to denoise the categorical data. We construct the posterior distribution of the categorical model as
\begin{equation}
    p_{\theta}(\bm{y}_{t-1} | \bm{y}_t, \bm{y}_0) = \mathcal{C}(\bm{y}_{t-1} | \bm{\theta}_{\text{post}}(\bm{y}_t, \bm{y}_0)),
\end{equation}
where 
\begin{equation}
    \bm{\theta}_{\text{post}}(\bm{y}_t, \bm{y}_0) = \bar{\bm{\theta}}/\sum_{k=1}^K\hat{\theta}_k
\end{equation}
and
\begin{equation}
    \bar{\bm{\theta}} = [\alpha_t \bm{y}_t + (1 - \alpha_t)/K] \odot [\hat{\alpha}_{t-1} \bm{y}_0 + (1 - \hat{\alpha}_{t-1})/K].
\end{equation}

Notably, as $\bm{y}_0$ is not known during inference, it must be estimated using a neural network. Inspired by DiffuSeq \cite{gong2022diffuseq, gong2023diffuseq}, we concatenate a representation of the source sequence to the noised representation of the target sequence to use as input to the $\bm{y}_0$ approximation model. In contrast to DiffuSeq, which uses the embedded representation of the source sequence $\bm{x}_0$, we process the source sequence with a transformer encoder model prior to concatenating, which is used as the memory component in the decoder architecture \cite{lewis2019bart}. Thus, the reverse diffusion process is fully modeled as
\begin{equation}
    p_{\theta}(\bm{y}_{t-1} | \bm{y}_t, \bm{x}_0) = \mathcal{C}(\bm{y}_{t-1} | \bm{\theta}_{\text{post}}(\bm{y}_t, \mathcal{F}_{\text{decoder}}(\bm{y}_t || \mathcal{F}_{\text{encoder}}(\bm{x}_0)))),
\end{equation}
where $||$ is the concatenation operator. We denote the estimated target sequence as $\hat{\bm{y}}_0$. Additionally, we add a sinusoidal embedding of the current diffusion timestep $t$ to the diffusion model input $\bm{y}_{t}$.

\subsection{Length Prediction}
\label{sec:length_prediction}

One downside of the diffusion methodology compared to normal autoregressive methods is the requirement that the total length of the sequence must be known prior to inference so that the starting uniform noise distribution can be properly initialized. Prior work, such as DiffuSeq \cite{gong2022diffuseq, gong2023diffuseq}, have primarily approached this issue by padding the sequence with observable \texttt{PAD} tokens up to a maximum length, and allowing the diffusion model to predict the location of \texttt{PAD} tokens as necessary. In contrast, MaskPredict \cite{ghazvininejad2019mask} presents the idea of a length prediction token, which is learnt by the transformer encoder block and used to initialize the input to the transformer decoder. We similarly prepend a \texttt{LENGTH} token to the source sequence $\bm{x}_0$ and use the learned representation of the \texttt{LENGTH} token for the sequence to predict a target length $\hat{\ell}_y$:
\begin{equation}
    p(\hat{\ell}_y|\bm{x}_0) = \mathcal{F}_{\text{length}}(\mathcal{F}_\text{encoder}(\bm{x}_0)),
\end{equation}
where $\mathcal{F}_{\text{length}}$ is a feed-forward classifier function over integers up to the maximum sequence length dependent only on the first column of $\mathcal{F}_\text{encoder}(\bm{x}_0)$.

Upon initial testing, we find that the baseline implementation of MaskPredict length prediction performs poorly for this task due to its rigidity in predicting the sequence length and frequency of predicting shorter sequences than the ground truth. To remedy this, we propose a novel adjustment to the method by appending a uniform random number $n \sim \mathcal{U}(1, N)$ of observable padding tokens to the decoder sequences during training. This increases the expected length of the sequence by $\frac{(1 + N)}{2}$ while allowing the diffusion model to predict pad tokens within a range of the the target length. Notably, as the probability of predicting $n$ padding tokens is equivalent for $n \in [1, N]$, the probability of predicting sequences with lengths within $\frac{(1 + N)}{2}$ of the predicted length is approximately equal. This offers benefits of both the MaskPredict and DiffuSeq approaches: The diffusion model is provided with information about the target length from the encoder while still allowing for the possibility of predicting a variety of sequence lengths rather than only the predicted target length. Intuitively, lower limits of $N$ place more focus on the length prediction component, while higher values of $N$ allow the model to vary more from the initial length prediction. Additionally, as the lengths of the input source sequence and output target sequence of the retrosynthesis task are highly correlated, we shift the methodology to predict the difference in length between the source and target rather than the total length of the target.

\subsection{Ensemble Voting of Models}

In order to stabilize and diversify the output of the diffusion model, we implement an ensemble of various individual diffusion models with different values of the random padding limit $N$. During inference, multiple samples are drawn from each model trained on different $N$. Outputs are then ranked according to the number of times they are sampled overall. Ties are broken using a ranked-choice voting scheme. This process encourages output diversity by combining models trained with varying degrees of reliance on the length prediction component. We refer to the constructed ensemble model as \ours.

We identically construct and train the individual models that compose \ours,  varying only the maximum number of random padding that can be added on to the target sequence during training. We test various voting combinations of models with random padding limit $N \in \{5, 10, 15, 20, 30, 40, 50, 60, 70, 80, 90\}$. We find that models trained with $N < 20$ perform poorly, as they place too much focus on the length prediction component, which is biased toward a lower increase in length. We construct the final ensemble using eight models trained on $N \in \{20, 30, 40, 50, 60, 70, 80, 90\}$.

\subsection{Loss Functions}

We utilize a combination of mean square error (MSE) and variational lower bound (VLB) losses to train the diffusion models, the latter of which is derived using the Kullback-Leibler (KL) divergence according to Hoogeboom et al \cite{hoogeboom2021argmax}. We apply MSE loss directly to the predicted target $\bm{y}_0$, while the VLB loss uses the sampled posterior of the predicted target:
\begin{align}
    \mathcal{L}_{\text{MSE}} &= \mathbb{E}\left[||\bm{y}_0^2 - \hat{\bm{y}}_0^2||\right] \\
    \mathcal{L}_{\text{VLB}} &= \mathbb{E}\left[\sum_{k=0}^K\bm{y}_{0, k} \log \hat{\bm{y}}_{0, k} - \sum_{t=2}^T\text{KL}(q(\bm{y}_{t-1}| \bm{y}_t, \bm{y}_0) | q(\bm{y}_{t-1}| \bm{y}_t, \hat{\bm{y}}_0))\right].
\end{align}
Finally, we include a loss term for the length prediction task using cross entropy:
\begin{equation}
    \mathcal{L}_\ell = \mathbb{E}\left[\sum_{l=0}^L{\ell_y}_l\log p(\hat{\ell_y}|\bm{x}_0)_l\right],
\end{equation}
where $\ell_y$ is a one-hot encoded vector representing the length difference between $\bm{y}_{0}$ and $\bm{x}_0$. Additionally, when training the diffusion models, we employ the importance-based time sampling algorithm used in DiffuSeq \cite{gong2022diffuseq, gong2023diffuseq}.

\subsection{Reproducibility}

We apply our models to the USPTO-50K \cite{Lowe2017} data set with Root-aligned SMILES augmentation \cite{zhong2022root}, which greatly improves the performance of the diffusion models. We directly utilize the dataset splits provided by the authors of Root-aligned SMILES. We form \ours as an ensemble of models with random padding limit $N \in \{20, 30, 40, 50, 60, 70, 80, 90\}$. Each model takes the form of a encoder-decoder transformer architecture with 6 layers and 8 attention heads with hidden dimension size 512, feed-forward size 2048, and GELU activation \cite{hendrycks2016gaussian}. We utilize an Adam optimizer with learning rate $1\times 10^{-4}$ and dropout rate of $0.1$. We set the number of diffusion steps to $T=200$. We use a cosine beta schedule \cite{chen2023importance} and a Gumbel-softmax distribution for noise sampling during the diffusion process.

During inference, we follow the work of Zhong et al. \cite{zhong2022root} to augment the input SMILES strings. 20 random Root-aligned SMILES strings are generated for each input product and provided to \ours for sampling. Thus, each model in \ours outputs 20 random samples for the input string. Output samples are canonized and ranked according to rate of occurrence.

\section{Related Work}
\label{sec:related}

\subsection{Retrosynthesis Models}

Early models for retrosynthesis were primarily rule-based, utilizing known scientific knowledge and synthetic reactions to construct algorithms to find retrosynthetic pathways \cite{cook2012computer, corey1985computer, sun2022computer}. Over time, these rule-based algorithms gave way to more data-driven approaches, which instead learn models based on known reactions and are more efficient when the number of reaction types is higher. Data-driven models include template-based, semi-template, and template-free generative models. In template-based approaches, models match features of a target product molecule to known reaction mechanisms in order to rank possible reactants \cite{coley2017computer, segler2017neural}, while semi-template methods first predict reaction centers and then complete the resulting synthons (incomplete molecules) to form possible reactants \cite{chen2023g, somnath2021learning}. In contrast, template-free generative models learn to construct new reactants, commonly using transformers \cite{zhong2022root, irwin2022chemformer, tu2022permutation} or Graph Neural Networks \cite{tu2022permutation, sacha2021molecule}. Here, generative models are sometimes considered to have the advantage, as they are better able to generalize to reactions that may exist outside the set of reaction templates, at the cost of losing scientific interpretability and trust.

Early template-free models took advantage of sequential models commonly used in natural language modeling. Liu et al. \citep{liu2017retrosynthetic} implement a sequence-to-sequence recurrent encoder-decoder model for retrosynthesis which operates on the SMILES \citep{weininger1988smiles} encodings of molecules, and demonstrate model efficacy on-par with rule-based mechanisms for retrosynthesis. In contrast to Liu et al.'s end-to-end approach, Segler et al. \citep{segler2017neural} adopt a neuro-symbolic approach, combining rule-based mechanisms with neural networks designed to predict which rules should be applied for the retrosynthesis. More recent approaches leverage further advances in natural language processing, directly using transformer architectures and training styles that have performed well in natural language and translation tasks. 

Irwin et al. introduce Chemformer \citep{irwin2022chemformer}, a transformer architecture similar to that of BART \cite{lewis2019bart}, a widely-used transformer in NLP for both sequence-to-sequence and discriminative learning tasks. Chemformer is pretrained in a self-supervised manner on SMILES strings for feature extraction, and then later applied to downstream tasks such as forward synthesis, property prediction, and retrosynthesis. Tu et al. \cite{tu2022permutation} leverage the permutation-invariance of graph neural network encoders to constuct a graph-to-sequence model, utilizing additional chemical features extracted by RDKit \cite{landrum2013rdkit}. Once the encoded representation of the graph is obtained, it is passed to a decoder model \cite{lewis2019bart} in order to produce a sequential SMILES output. Finally, Zhong et al. \cite{zhong2022root} modify the mechanisms by which SMILES strings are constructed for molecules in order to better represent relationships between products and reactants before they are passed to neural machine translation models. Their root-aligned SMILES methodology is able to significantly improve upon state-of-the-art baselines while applying only pre-existing language models.

An important aspect of recent works which focus on encoding and decoding SMILES representations of molecules is the transformer architecture. Proposed in 2017 for natural language tasks \cite{vaswani2017attention}, transformer architectures are highly efficient, offering good parallelization while achieving strong performance on sequential input. Recent analysis of transformers have compared their inner workings favorably to graph learning algorithms \cite{joshi_2020}, offering some insight as to why they are able to perform strongly on chemistry-related tasks when applied to SMILES strings.

However, one downside of encoder-decoder transformer models lies in their auto-regressive nature. When predicting SMILES outputs, transformer models are generally required to do so token-by-token, re-using previous outputs as the input for the next prediction in the sequence. This requires a significant amount of planning on the part of the transformer, which is unable to amend previous predictions when prediction later parts of the sequence. Additionally, chemical molecules are most often non-sequential themselves, often containing both ring structures and branches. Thus, we hypothesize that methods which decode the entire sequence at once rather than token-by-token may achieve greater success when it comes to predicting molecules.

\subsection{Categorical Diffusion Models}
Diffusion models are a class of generative model that produce high quality results from uniform random noise. In the forward pass, random noise is iteratively added to a target sample over $T$ steps until the sample is indistinguishable from random noise. In the backward pass, the model is trained to iteratively de-noise the random sample until the target sample is reconstructed \cite{sohl2015deep, ho2020denoising, nichol2021improved}. Additionally, conditional diffusion models \cite{saharia2022photorealistic, gu2022vector, zhang2023adding} allow for the inclusion of conditional constraints and properties during the de-noising process (often in the form of textual information), helping guide the model toward a target output with specific parameters or properties. 

Traditional diffusion models operate over continuous regimes such as image generation \cite{sohl2015deep, ho2020denoising}, hindering their application on discrete problems such as language or molecular generation tasks. To extend these models to discrete domains, many adaptations have been proposed, such as multinomial diffusion \cite{hoogeboom2021argmax} and DiffuSeq \cite{gong2022diffuseq, gong2023diffuseq}. DiffuSeq adapts conditional diffusion models for the task of dialogue generation by employing traditional diffusion within the token embedding space. The authors concatenate embedded representations of the target embeddings and the source embeddings, which act as the conditional. Gaussian noise is then applied to the portion of the sample containing the target embeddings, and the source embeddings are left untouched during the diffusion process. Yuan et al. \cite{yuan2022seqdiffuseq} follow the work of DiffuSeq, but utilize alternative network architectures and support adaptive noise scheduling. Dieleman et al. \cite{dieleman2022continuous} similarly add noise after input embeddings are applied, and jointly learn both the diffusion and embedding model. In contrast, the multinomial diffusion proposed by Hoogeboom et al. \cite{hoogeboom2021argmax} applies noising in the discrete domain, repeatedly sampling from a categorical distribution conditioned on the current diffusion timestep and the current sample. This approach is similarly employed by Austin et al. \cite{austin2021structured} and DiffusionBERT \cite{he2022diffusionbert}. Austin et al. liken the noise-adding approach of traditional diffusion to Gaussian transition matrices over discrete space, and are able to improve the performance of text diffusion models by properly choosing the transition matrix. DiffusionBERT takes an approach similar to MaskPredict \cite{ghazvininejad2019mask}, a parallel-decoding schema for text generation that operates similarly to diffusion models. Instead of sampling tokens from a categorical distribution when adding noise, MaskPredict and DiffusionBERT randomly mask-out tokens at varying rates during training, which are un-masked by the model. During inference, both methods generate sequences from a sequence of masked tokens, with no need to initialize a noise vector, which may limit diversity in the generated output.

\section{Results}
\label{sec:results}

\subsection{Overall Performance}

\begin{table}[h!]
    \small
    \centering
    \caption{Top-K accuracy for tempate-based, semi-template, and template-free retrosynthesis models. The best performing model of its category for each $K$ is 
    in \textbf{bold} and the second best model is \underline{underlined}. \ours is the best performing model for $K=1$ and the second best model for $K=3,5$ among template-free methods.}
    \begin{tabular}{l l r r r r}
        \toprule
         Category & Model & K=1 & 3 & 5 & 10 \\
        \midrule
        \multirow{4}{*}{Template-based} & Retrosim \cite{coley2017computer} & 37.3 & 54.7 & 63.3 & 74.1 \\
                                           & Neuralsym \cite{segler2017neural} & 44.4 & 65.3 & 72.4 & 78.9 \\
                                           & GLN \cite{dai2019retrosynthesis} & \underline{52.5} & \underline{69.0} & \underline{75.6} & \underline{83.7} \\
                                           & LocalRetro \cite{chen2021deep} & \textbf{53.4} & \textbf{77.5} & \textbf{85.9} & \textbf{92.4} \\
        \midrule
        \multirow{5}{*}{Semi-template} & G2Gs \cite{shi2020graph} & 48.9 & 67.6 & 72.5 & 75.5 \\
                                          & GraphRetro \cite{somnath2021learning} & 53.7 & 68.3 & 72.2 & 75.5 \\
                                          & RetroXpert \cite{yan2020retroxpert} & 50.4 & 61.1 & 62.3 & 63.4 \\
                                          & RetroPrime \cite{wang2021retroprime} & 51.4 & 70.8 & 74.0 & \underline{76.1} \\
                                          & G$^2$Retro \cite{chen2023g} & \underline{53.9} & \underline{74.6} & \underline{80.7} & \textbf{86.6} \\
                                          & GDiffRetro \cite{sun2025gdiffretro} & \textbf{58.9} & \textbf{79.1} & \textbf{81.9} & - \\
        \midrule
        \multirow{11}{*}{Template-free} & Seq2Seq \cite{liu2017retrosynthetic} & 37.4 & 52.4 & 57.0 & 61.7 \\
                                          & Levenshtein \cite{sumner2020levenshtein} & 41.5 & 48.1 & 50.0 & 51.4 \\
                                          & GTA \cite{seo2021gta} & 51.1 & 67.6 & 74.8 & 81.6 \\
                                          & Graph2SMILES \cite{tu2022permutation} & 51.2 & 66.3 & 70.4 & 73.9 \\
                                          & Dual-TF \cite{sun2021towards} & 53.3 & 69.7 & 73.0 & 75.0 \\
                                          & MEGAN \cite{sacha2021molecule} & 48.1 & 70.7 & 78.4 & 86.1\\
                                          & Chemformer \cite{irwin2022chemformer} & 54.3 & - & 62.3 & 63.0 \\
                                          & Retroformer \cite{wan2022retroformer} & 53.2 & 71.1 & 76.6 & 82.1 \\
                                          & Tied transformer \cite{kim2021valid} & 47.1 & 67.2 & 73.5 & 78.5 \\
                                          & R-SMILES \cite{zhong2022root} & \underline{56.3} & \textbf{79.2} & \textbf{86.2} & \textbf{91.0} \\
        \cmidrule(lr){2-6}                                   
                                          & \ours & \textbf{57.6} & \underline{79.0} & \underline{84.1} & \underline{87.4} \\
        \midrule
                                           
    \end{tabular}
    \label{tab:topk}
\end{table}

We compare \ours to a variety of template-based, semi-template, and template-free methods for retrosynthesis prediction with unknown reaction types. We report top-k accuracy for $k\in\{1,3,5,10\}$ following standard procedures. We construct \ours as an ensemble of eight diffusion models with various hyperparameters and sample from each model twenty times with augmented input SMILES according to the work of Zhong et al. \cite{zhong2022root}. Predicted reactants sampled from the diffusion models are ranked according to the number of times they were predicted by the diffusion models. By using multiple different models with different hyperparameters to predict possible reactants for the same reaction, we encourage diversity and stability in the output reactants. Ties in reactant rankings are decided according to a ranked choice voting scheme. Further ties are broken arbitrarily. Results for the diffusion ensemble model are reported against baseline algorithms in Table \ref{tab:topk}, while individual model results are reported in Table \ref{tab:our_models}. We include a short review of the baselines used for comparison in appendix \ref{sec:A1}.

\begin{figure}[h]
    \small
    \centering
    \includegraphics[width=0.7\linewidth]{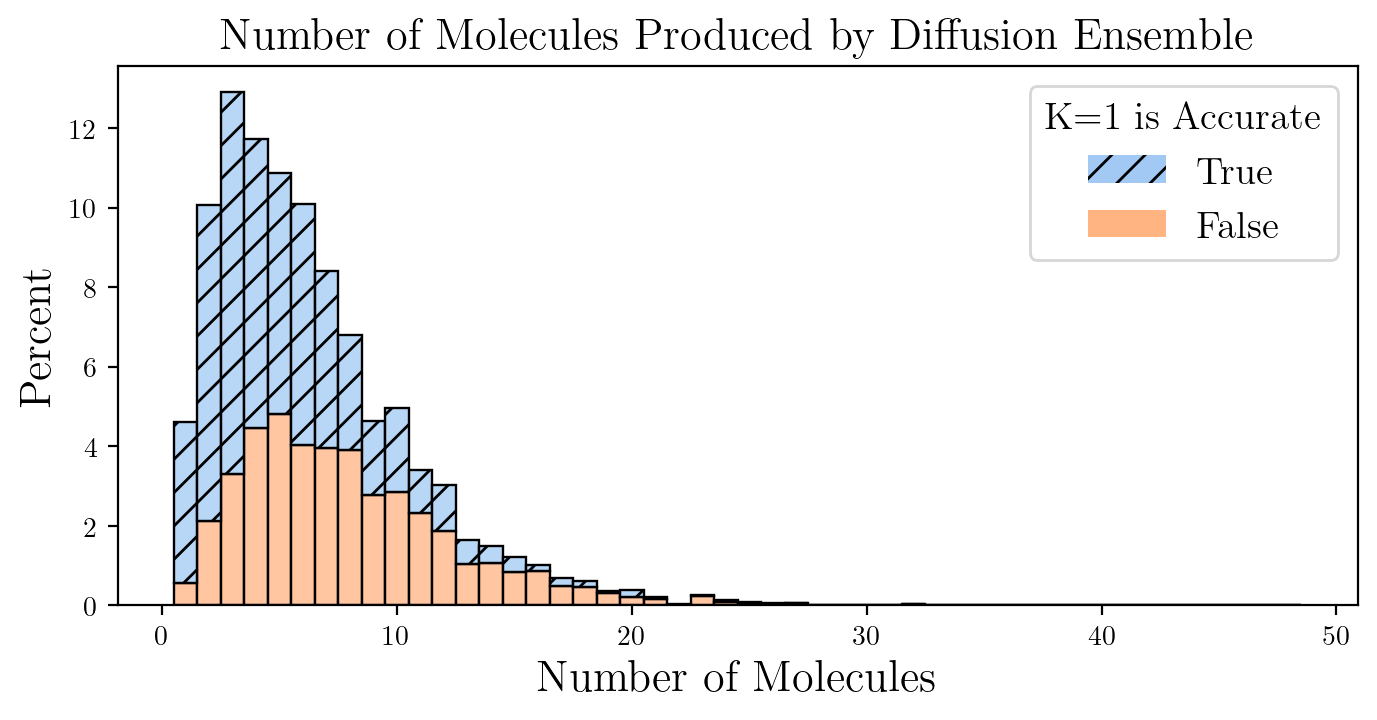}
    \caption{A histogram of the number of different molecules output by \ours for each reaction during inference.The distribution has a mean of 10.0 and a median of 9. ``K=1 is Accurate'' indicates if the most commonly predicted molecule matched the ground-truth reaction.}
    \label{fig:num_samples}
\end{figure}

Notably, \ours is the best performing model for $K=1$ (57.6\%) across all template-free models and is highly competitive for higher values of $K$, achieving the second best performance for $K=3$, $K=5$, and $K=10$ (79.0\%, 84.1\%, and 87.4\%, respectively) for template-free methods behind Root-Aligned SMILES. Comparing to the state-of-the-art $K=1$ result of GDiffRetro \cite{sun2025gdiffretro}, \ours achieves comparable $K=3$ performance and out-performs GDiffRetro at higher values of $K$, indicating \ours produces a wider array of possible reactants. Notably, compared to R-SMILES, both \ours and GDiffRetro struggle with output diversity, resulting in both diffusion-based models having decreased performance at high $K$. The decrease in performance for these models at high $K$ can likely be attributed to the affinity for diffusion models to most often sample from the peak of the posterior distribution: despite generating outputs from multiple models for a diversity of input product SMILES, the output of \ours tends to produce just a few candidate reactant sets for each product, producing the same output molecule numerous times in a different, non-canonical SMILES form. Indeed, Figure \ref{fig:num_samples} shows a histogram of the number of sample reactions generated for the test dataset. The diffusion ensemble produces a median number of samples of 5, with 17.9\% of reactions producing less than 5 samples, and 57.1\% producing less than 10. Notably, we find that fewer reactants tend to be sampled when \ours is correct in its output: reactions where the top-1 reactant is correctly predicted produce only 8.4 different molecules on average, compared to 12.3 for cases where the top-1 reactant does not match the ground truth. This lack of outputs is less observed in other methods such as R-SMILES which apply beam-search algorithms to sequential decoders, allowing a greater number of possible reactants to be found. 

We further analyze the performance of \ours across varying reaction types, as presented in Table \ref{tab:rxntypes}. Notably, the performance of \ours is inconsistent among reaction types, which has also been observed in other models \citep{chen2023g}. The diffusion ensemble performs strongly on reactions which operate on ring structures, such as acyclation, but performs poorly at predicting patented reactants for ring formation, through either heterocycle formation or C-C bond formation. Compared to G$^2$Retro \citep{chen2023g}, the best performing semi-template method which offers a similar analysis of their model, \ours is significantly more capable of accurately predicting the reactants for protection reactions
\cite{chen2023g}. For other reaction types, the overall trends in performance are rather similar for both \ours and G$^2$Retro. Both models exhibit their best performance on functional group addition followed by acyclation, and their worst performance on heterocycle formation, functional group interconversion, and C-C bond formation.

\begin{table}[h]
    \small
    \centering
    \caption{Top-K accuracy for \ours grouped by reaction type. Rows are ordered by number of reactions of the specified reaction type in descending order.}
    \begin{tabular}{l r r r r r}
        \toprule
         Reaction Type & K=1 & 3 & 5 & 10 & Support\\
        \midrule
        Heteroatom alkylation and arylation & 58.6 & 81.1 & 86.0 & 89.3 & 1,516 \\
        Acyclation and related processes & 70.4 & 88.7 & 92.2 & 93.9 & 1,190 \\
        Deprotections & 53.4 & 79.8 & 84.9 & 87.2 & 822 \\
        C-C bond formation & 42.2 & 64.2 & 72.0 & 77.8 & 567 \\
        Reductions & 57.2 & 75.2 & 81.4 & 86.2 & 463 \\
        Functional group interconversion & 39.0 & 58.2 & 66.5 & 73.1 & 182 \\
        Heterocycle formation & 33.3 & 48.9 & 55.6 & 60.0 & 90 \\
        Oxidations & 60.2 & 84.3 & 88.0 & 91.6 & 83 \\
        Protections & 64.2 & 83.6 & 91.0 & 91.0 & 67 \\
        Functional group addition & 78.3 & 87.0 & 87.0 & 87.0 & 23 \\
        \midrule
    \end{tabular}
    \label{tab:rxntypes}
\end{table}

\subsection{Individual Model Performance}

We construct \ours from eight individually trained diffusion models, each of which uses a different random padding limit $N$ (described in Section \ref{sec:length_prediction}) during training. We report the individual performance for each of the trained models in Table \ref{tab:our_models}. We compare these results to the \ours ensemble as well as a baseline diffusion model trained using direct length prediction, without the length variation used in the individual \ours models, also reported in Table \ref{tab:our_models}.

The individual \ours models greatly improve upon the baseline length prediction model, demonstrating the effectiveness of our length variation technique. The length variation technique encourages the individual models to over-predict the length of the reactant SMILES and then reduce the output SMILES to the correct size using predictable padding tokens. Meanwhile, the baseline length prediction model drastically under-performs all other models in terms of accuracy, demonstrating the sensitivity of the diffusion models for SMILES string generation to the size of the noise vector. Both the individual \ours models and the baseline length prediction model exhibit low variance in the output reactants, only predicting 3.2 and 3.8 reactants on average, respectively. Notably, all models still achieves high validity from among the output molecules, indicating that the individual diffusion models are properly learning not only the SMILES grammar, but also rules of molecular validity.

When the individual models are ensembled into the complete \ours model, both the accuracy and diversity of predicted reactants are significantly improved. The \ours ensemble produces on average twice the number of reactants compared to the individual models, highlighting the impact of the different individual model setups on the output molecules. By combining multiple models trained on different upper limits for the length of the SMILES string, we can leverage different weightings of the length prediction mechanism. When a model is trained on a lower padding limit, more emphasis is placed on the length prediction component, as the model is less able to diverge from the predicted length. When the padding limit is larger, the diffusion model has more freedom to construct SMILES of varying lengths, but are less informed by the length prediction. By combining multiple models on various random padding limits, we improve the diversity and accuracy of the predicted reactants.

\begin{table}[h]
    \small
    \centering
    \caption{Comparison of the \ours ensemble with the individual models that make up \ours. Individual models are notated as \ours$_N$, where $N$ represents the upper bound on the number of added padding tokens according to the procedure in Section \ref{sec:length_prediction}. We additionally include single models with baseline length prediction and oracle length prediction. The individual \ours models show significantly higher accuracy than baseline length prediction, but are outperformed by the \ours ensemble. The oracle length model drastically outperforms all existing models.}
    \begin{tabular}{r r r r r r r}
        \toprule
        Model & K=1 & 3 & 5 & 10 & Sample Validity & Avg. Num. Reactants \\
        \midrule
        \ours Ensemble & 57.6 & 79.0 & 84.1 & 87.4 & 100.0 & 10.0 \\
        \midrule
        \ours$_{20}$ & 53.2 & 70.3 & 72.9 & 73.6 & 100.0 & 3.2 \\
        \ours$_{30}$ & 55.2 & 71.5 & 74.4 & 75.2 & 99.9 & 3.2 \\
        \ours$_{40}$ & 54.6 & 72.1 & 74.4 & 75.3 & 99.9 & 3.2 \\
        \ours$_{50}$ & 54.9 & 71.3 & 73.7 & 74.4 & 100.0 & 3.2 \\
        \ours$_{60}$ & 55.4 & 71.7 & 74.6 & 75.4 & 99.9 & 3.3 \\
        \ours$_{70}$ & 54.3 & 71.1 & 73.5 & 74.4 & 99.6 & 3.2 \\
        \ours$_{80}$ & 54.6 & 71.9 & 74.4 & 75.1 & 99.6 & 3.3 \\
        \ours$_{90}$ & 54.5 & 71.6 & 74.2 & 74.9 & 99.8 & 3.3 \\
        \midrule
        Baseline Length & 40.4 & 55.9 & 58.8 & 59.9 & 99.9 & 3.8 \\
        Oracle Length & 77.0 & 88.1 & 89.5 & 90.0 & 99.7 & 2.8 \\
        \midrule
    \end{tabular}
    \label{tab:our_models}
\end{table}

\subsection{Upper Limits on Performance}

In this section, we consider a upper limit on the performance of categorical diffusion models for chemical retrosynthesis by considering an oracle model which predicts the length of the output SMILES string with 100\% accuracy. This allows us to construct a diffusion model without needing to incorporate aspects of length prediction or variability in the size of the output as discussed in Section \ref{sec:length_prediction}. Under the assumption of the length-predicting oracle, we can initialize input noise to the diffusion model of the proper size. We run this experiment for a single diffusion model and utilize the same parameters and repeated sampling during inference as discussed in the experimental setup.

Results for this experiment are presented in Table \ref{tab:our_models} in comparison to the \ours models and the baseline length prediction diffusion model. The model with oracle length drastically outperforms \ours as well as all existing methods for $K=1,3,5$, but performs slightly worse than both Root-aligned SMILES and LocalRetro for $K=10$. Similarly to \ours, this can be attributed to a lack of variety in the predicted output, which is even more extreme for the oracle length model: on average, only 2.8 different reactants are predicted, with a median of just 2 predicted reactants. However, this lack of diversity is overcome by the oracle model's high accuracy amongst the predicted reactants.

This result highlights the importance of accurate length prediction in non-autoregressive models, particularly in the case of molecular generation with SMILES strings. While there is some variability in the length of randomly generated SMILES string for a specific molecule, these variants generally only differ by a length of two or three, if they differ at all. If the length prediction is off from a viable value by even one or two, it may force an entirely different molecule to be generated, resulting in lower reported performance. By including the random length padding in \ours, we help combat the diffusion model's sensitivity to length prediction, but remain far from the performance of a model with perfect length prediction.

\subsection{Case Studies}
\label{sec:case_study}

\begin{figure}[!h]
    \footnotesize
    \centering
    \begin{tabular}{c c c c c}
        \multirow{2}{*}{\textbf{a)}} &
        \subf{\raisebox{0.3\height}{\includegraphics[width=0.15\linewidth]{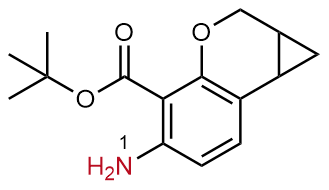}}}{\textbf{3aa}, desired product} & 
        \subf{\includegraphics[width=0.15\linewidth]{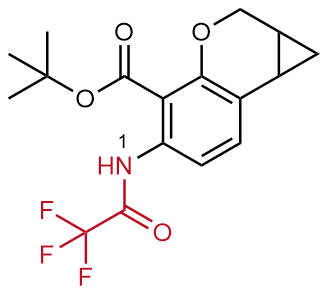}}{\textbf{3ab}, ground-truth reactants} & \\ &
        \subf{\includegraphics[width=0.15\linewidth]{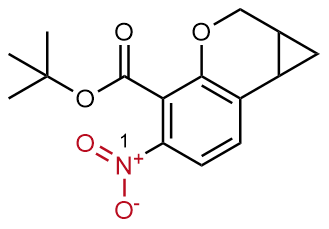}}{\textbf{3ac}, top-1 reactants (89.4\%)} & 
        \subf{\includegraphics[width=0.15\linewidth]{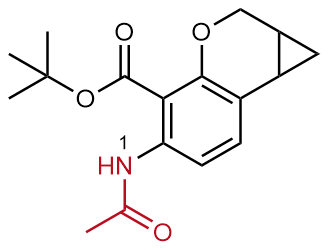}}{\textbf{3ad}, top-2 reactants (3.0\%)} & 
        \subf{\includegraphics[width=0.15\linewidth]{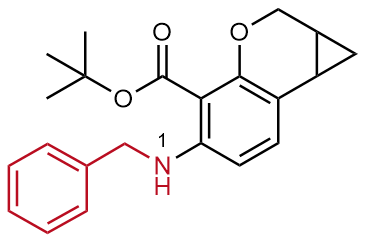}}{\textbf{3ae}, top-3 reactants (1.5\%)} \\
        \\
        \multirow{2}{*}{\textbf{b)}} &
        \subf{\includegraphics[width=0.18\linewidth]{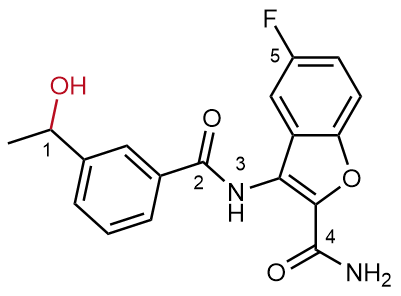}}{\textbf{3ba}, desired product} & 
        \subf{\includegraphics[width=0.18\linewidth]{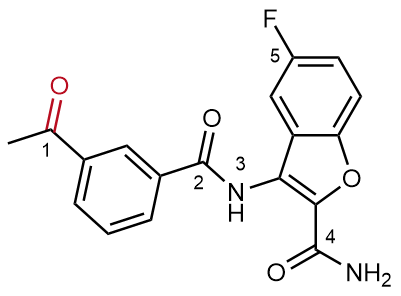}}{\textbf{3bb}, ground-truth reactants} & \\ &
        \subf{\includegraphics[width=0.18\linewidth]{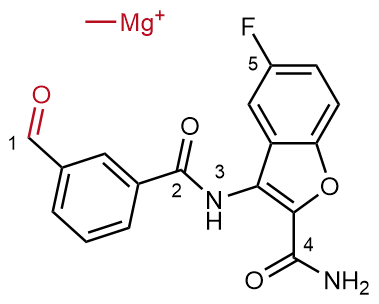}}{\textbf{3bc}, top-1 reactants (42.2\%)} & 
        \subf{\includegraphics[width=0.18\linewidth]{images/molecules/case12_target.png}}{\textbf{3bd}, top-2 reactants (38.3\%)} & 
        \subf{\includegraphics[width=0.24\linewidth]{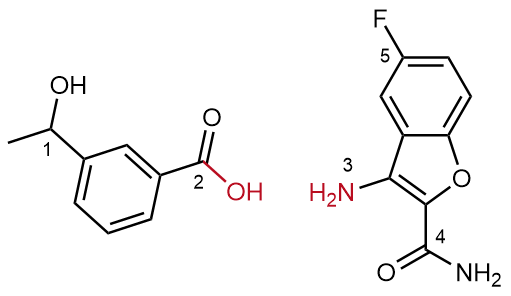}}{\textbf{3be}, top-3 reactants (9.7\%)} \\
        \\
        \multirow{2}{*}{\textbf{c)}} &
        \subf{\includegraphics[width=0.24\linewidth]{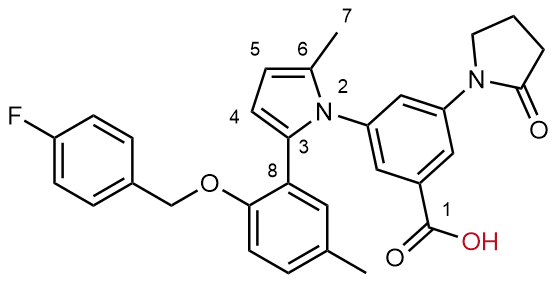}}{\textbf{3ca}, desired product} & 
        \subf{\includegraphics[width=0.24\linewidth]{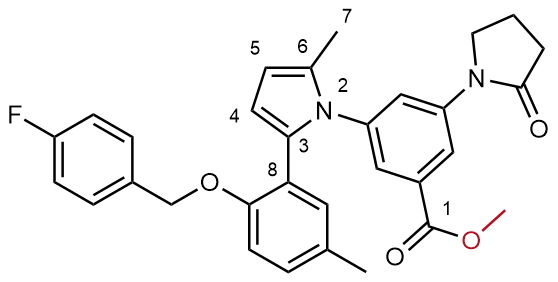}}{\textbf{3cb}, ground-truth reactants} & \\ &
        \subf{\includegraphics[width=0.24\linewidth]{images/molecules/case17_target.png}}{\textbf{3cc}, top-1 reactants (61.3\%)} & 
        \subf{\includegraphics[width=0.24\linewidth]{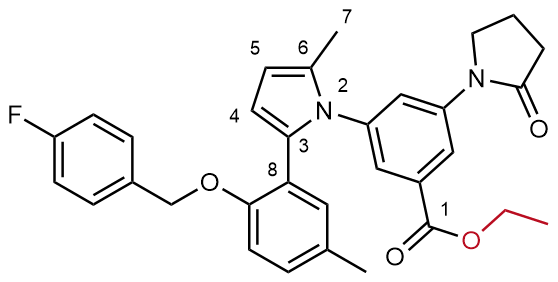}}{\textbf{3cd}, top-2 reactants (18.7\%)} & 
        \subf{\includegraphics[width=0.29\linewidth]{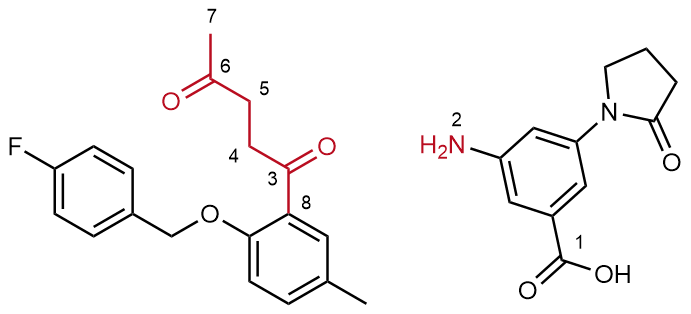}}{\textbf{3ce}, top-3 reactants (16.7\%)} \\
        \multirow{2}{*}{\textbf{d)}} &
        \subf{\includegraphics[width=0.2\linewidth]{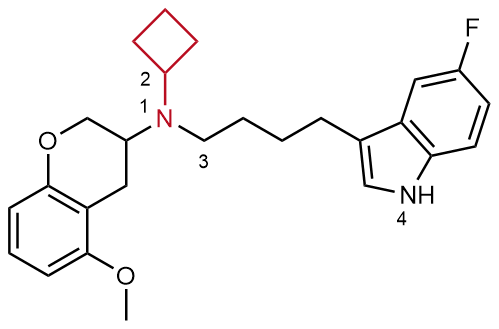}}{\textbf{3da}, desired product} & 
        \subf{\includegraphics[width=0.2\linewidth]{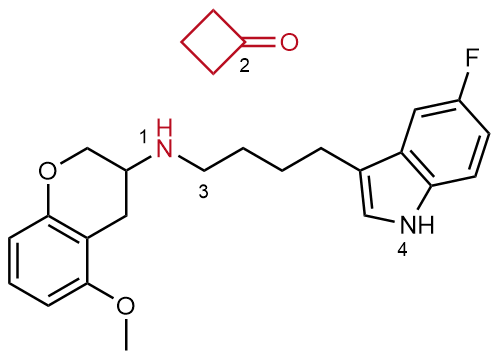}}{\textbf{3db}, ground-truth reactants} & \\ &
        \subf{\includegraphics[width=0.2\linewidth]{images/molecules/case18_target.png}}{\textbf{3dc}, top-1 reactants (74.3\%)} & 
        \subf{\includegraphics[width=0.24\linewidth]{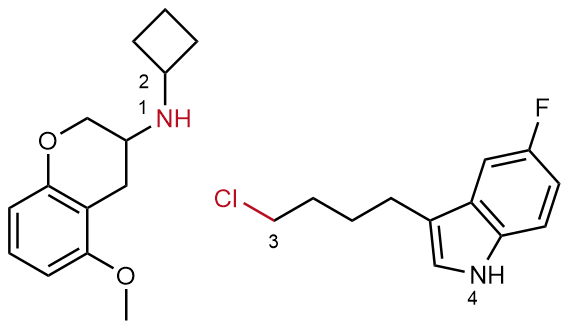}}{\textbf{3dd}, top-2 reactants (11.5\%)} & 
        \subf{\includegraphics[width=0.24\linewidth]{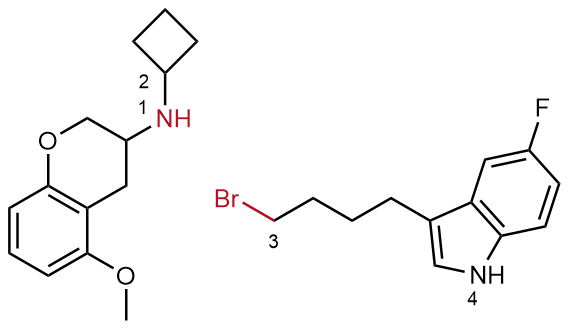}}{\textbf{3de}, top-3 reactants (2.7\%)} \\
    \end{tabular}
    \caption{Top-3 reactants for 4 different target products from the test set compared to patent reactants. Values in parentheses indicated the percentage of time that reactant was sampled from the diffusion models. Numbers next to atoms are labels used to refer to the atom. Colored regions indicate where in the molecule a reaction is expected to occur.}
    \label{fig:case_study}
\end{figure}

Our final results are in the form of a case study of four different reactions, the results of which can be seen in Figure \ref{fig:case_study}. We select four different reactions from the test dataset to highlight the successes and limitations of the diffusion models. For each reaction, we report the top-3 reactants sampled from the diffusion ensemble.

Reaction \textbf{a} (Fig. \ref{fig:case_study}a) is a deprotection reaction in the forward direction in which the trifluoroacetamide protecting group (Fig. \ref{fig:case_study}ab) is reacted with sodium borohydride in an ethanol solution to produce a primary amine at at the N1 position (Fig. \ref{fig:case_study}aa). The diffusion ensemble does not predict this reactant, however, instead opting for the reduction of a nitro group, predicted 89.4\% of the time, (Fig. \ref{fig:case_study}ac) at the N1 position. This is a widely used and viable reaction involving metal catalyzed reduction of the nitro group using Fe, Zn, Pd or Ni. Similarly, the third ranked choice (Fig. \ref{fig:case_study}ae) is also a viable option involving a debenzylation reaction which is commonly performed in the presence of a metal catalyst under a hydrogen atmosphere. The second choice reaction (Fig. \ref{fig:case_study}ad), however, undergoes amide hydrolysis under harsh conditions such as a strong acid. While this is also a viable synthesis, a harsh condition such as a strong acid would likely interact with the tert-butyl ester group leading to unwanted side products. Notably, the diffusion ensemble does not predict the patented reactant, primarily predicting the reduction of a nitro group 89.4\% of the time. We hypothesize that this is a result of the specific use-case of the ground-truth reaction, which is likely used in an industrial setting. Unlike the top-1 (Fig. \ref{fig:case_study}ac) reactant, the ground-truth (Fig. \ref{fig:case_study}ab) does not use the metal catalyzed reduction of a nitro group (top-1) or debenzylation (top-3) under a hydrogen atmosphere because it is not a safe reaction on industrial scale. While the top-1 or top-3 reactants may actually be more common in an academic lab setting, the ground-truth reactant is likely preferable on a large-scale industrial setting.

Reaction \textbf{b} (Fig. \ref{fig:case_study}b) is a common reduction reaction of a ketone at C1 (Fig. \ref{fig:case_study}bb) with sodium borohydride in a methanol solution to produce a secondary alcohol (Fig. \ref{fig:case_study}ba). This same reaction is also predicted by the diffusion models 38.3\% of the time to give the second most popular predicted reactants (Fig. \ref{fig:case_study}bd). The most popular predicted reactants  is an aldehyde and  Grignard reagent, predicted 42.2\% of the time, in which the methylmagnesium bromide reacts with the aldehyde at position C1 (Fig. \ref{fig:case_study}bc) to produce the desired product in the forward reaction. However, the Grignard reagent (MeMgBr), which is a more nucleophilic reagent, is likely to react with other functional groups present in the molecule as well, such as the two amides at C2 and C4 or displace the fluorine at C5 via a nucleophilic aromatic substitution-like type reaction. In contrast, while the ground-truth/second-most predicted reaction (Fig. \ref{fig:case_study}bb/\ref{fig:case_study}bd) could lead to unwanted products, such as the reduction the carbonyl groups at C2 and C4, these reactions are less likely to occur compared to the Grignard reaction under a proper choice of reagents such as sodium borohydride in ethanol. The top-3 reactant is predicted just 9.7\% of the time, and unlike the other proposed reactants, it does not interact with the C1 carbon. Instead, it disconnects the molecule at the C1-N3 amide to reveal a C1 carboxylic acid and an N3 amine.  (Fig. \ref{fig:case_study}be). The amide bond can be constructed in the forward reaction using a suitable amide coupling reaction condition. Yet again, this reaction is viable but would likely produce unwanted byproducts if other functional groups such as the free secondary alcohol or the primary amide in the proposed reaction were not protected first.

Reaction \textbf{c} (Fig. \ref{fig:case_study}c) is an ester hydrolysis in which the C1 ester (Fig. \ref{fig:case_study}cb) is reacted with sodium hydroxide and a water/ethanol mixture to produce a carboxylic acid (Fig. \ref{fig:case_study}ca). This reaction is also predicted by the diffusion ensemble a vast majority of the time, forming 61.3\% of the sampled reactants (Fig. \ref{fig:case_study}cc). The top-2 reactant is also a viable alternative to the ground-truth reactant, just with a ethyl ester at C1 instead of a methyl ester, which is also a common reaction. Interestingly, the top-3 reactant sampled by the diffusion models is a Paal-Knorr-type pyrrole synthesis \cite{paal1884ueber, knorr1884synthese} from an aniline (N2) and the diketone (C3-C6)(Fig. \ref{fig:case_study}cd). The Paal-Knorr reactions usually involves harsh conditions such as strong acids which may not be compatible with the ether or the lactam functional groups \cite{paal1884ueber, knorr1884synthese}. Also, the free C1 carboxylic acid may need to be protected as an ester before the pyrrole synthesis to produce the ground-truth reactant molecule (Fig. \ref{fig:case_study}cb) which would need to undergo the aforementioned ester hydrolysis to produce the target molecule. Thus, \ref{fig:case_study}cd could be viewed as one-step further back in the retrosynthesis chain compared to the ground-truth/top-1 reactant. 

Finally, reaction \textbf{d} (Fig. \ref{fig:case_study}d) is a Borch reductive amination \cite{borch2003reductive} between cyclobutanone and a secondary amine in the presence of a reducing agent (Fig. \ref{fig:case_study}db) to form a tertiary amine with a cyclobutyl group at N1 (Fig. \ref{fig:case_study}db). The ground-truth reactants are the top-predicted reactants by \ours, making up 74.3\% of the predicted reactants. Both the second and third most predicted reactants are amine alkylations involving halides, in which an alkyl halide at C3 (Fig. \ref{fig:case_study}dd/\ref{fig:case_study}de) reacts with the N1 amine to form a tertiary amine. Such N-alkylation  reactions are commonly used when reduction amination does not work, and are valid alternative reactions. However, potential side reactions such as N-1 overalkylation or intramolecular N-4 alkylation would likely lead to unwanted byproducts. As such, the ground-truth/top-1 reactants are preferred.

These case studies demonstrate that the diffusion models are capable of learning a variety of viable reactions for organic synthesis, but like many other models, struggle to understand greater domain requirements or the likelihood of byproducts, as exemplified in the top-1 reactant for reaction \textbf{b} (Fig. \ref{fig:case_study}bc) and top-2/3 reactants for reaction \textbf{d} (Fig. \ref{fig:case_study}dd/\ref{fig:case_study}de). However, the models also demonstrate significant capability to properly utilize common and routine reactions (Fig. \ref{fig:case_study}ac and Fig. \ref{fig:case_study}cc) as well as more complex reactions such as ring formations (Fig. \ref{fig:case_study}ce). Even in cases where the ground-truth reactants were not predicted in the top-3 reactants, the diffusion models produced suitable alternatives used in other reactions (Fig. \ref{fig:case_study}ac) to accomplish the same goal. This highlights a key difficulty in assessing the accuracy and performance of models for retrosynthesis; oftentimes, multiple different reactions may be viable, and the use case for each reaction may depend on information not present in the molecular structure, such as the case of reaction \textbf{a}. Metrics of ground-truth accuracy do not capture the viability of a reaction, just that they match the patented reaction, which may lead to misrepresentation of the model's capabilities when a suitable alternative is predicted.

\section{Discussions and Conclusions}
\label{sec:discussion}

Our results show that categorical diffusion models offer a competitive alternative to auto-regressive models for template-free single step retrosynthesis prediction and are able to outperform many state-of-the-art template-based, semi-template, and template-free methods on top-1 accuracy, and achieve similar performance to state-of-the-art models on top-3, top-5, and top-10 accuracy. The proposed \ours ensemble is able to efficiently sample from the posterior distribution of possible reactant SMILES strings while maintaining the viability of output SMILES. \ours is capable of reproducing existing patented reactions as well as proposing new and viable reactants (Section \ref{sec:case_study}).

\subsection{Comparison to Other Template-free Methods}

Compared to other template-free methods, our diffusion ensemble is able to construct the entire reactant SMILES string in unison, rather than token by token. Methods like Chemformer \cite{irwin2022chemformer} or R-SMILES \cite{zhong2022root} build the SMILES sequence by predicting a distribution of probabilities for the next token conditioned on the current sequence and then selecting the token with the highest probability and adding it to the sequence. To generate multiple possible reactants, auto-regressive methods generally use beam-search algorithms to generate multiple possible reactants \cite{irwin2022chemformer}. Beam-search algorithms explore multiple options at each step, and only continue exploring the top-k sequences with the highest probability. In contrast, our diffusion approach directly samples an entire reactant SMILES string from the approximated posterior distribution of possible reactants. Thus, the proportion of times a molecule is sampled from \ours can be interpreted as a approximation of the posterior probability of that reactant. This offers an interpretable measure of confidence in the model outputs. However, this form of sampling also results in limited diversity in the output reactant set: when the model is highly confident or little diversity exists in the space of possible reactants, only a few possible reactants may be produced. In contrast, when the model is unsure or there is a high number of possible reactions, more reactants will be produced. This trade-off can be disadvantageous in certain circumstances, particularly when there is a low number of predicted reactants and the top-predicted reactant is unsuitable.

\subsection{Comparison to Template-based and Semi-template Methods}

Unlike template-based and semi-template methods, diffusion models do not directly utilize a static set of reaction templates. Rather, \ours learn suitable reactions directly from the data, allowing for greater generalization to reactions that may occur outside the realm of existing templates. However, this comes at a notable cost in interpretability: because the diffusion method constructs the proposed reactants from randomly generated uniform noise, it is not always clear what causes the emergence of a particular reactant over another possible reactant. A greater analysis of the latent noise would need to be done to further understand the workings of \ours. In contrast, template-based and semi-template based methods directly utilize the existing molecular structure and set of reaction templates when determining where reaction centers are and which reaction templates apply. This offers a greater level of interpretability and scientific backing than template-free methods.

\subsection{Limitations and Directions of Future Work}

Despite the strong performance of \ours, there are a few notable limitations to the modeling approach. Firstly, raw diffusion models are highly sensitive to the predicted size of the output. Thus, methods such as our novel variant length padding (Section \ref{sec:length_prediction}) must be employed to diversify target lengths and allow diffusion models to predict molecules of varying size. Secondly, \ours suffers from a lack of output variety, often producing just a few possible reactants. Because diffusion models sample from an approximated posterior, reactants which the model deems more probable will be sampled at a higher rate, resulting in fewer reactants being sampled overall for a static sample size. This is particularly true in cases where the model is confident in its prediction. Finally, the model sometimes produces sub-optimal reactions, such as some of those displayed in Figure \ref{fig:case_study}. Such reactions often undergo additional reactions alongside the target reaction, and while the desired product is a possible outcome, additional byproducts would likely be produced in many cases. These limitations highlight necessary areas of future research for not only categorical diffusion models in chemical retrosynthesis, but also categorical diffusion models for sequence generation in general:
\begin{enumerate}
    \item Categorical diffusion models must be able to adapt to differently sized outputs. Prior work relies on predictable padding tokens to predict differently sized sequences \cite{gong2022diffuseq, gong2023diffuseq}, but this reliance can introduce its own biases due to the prevalence of padding tokens for shorter sequences. \ours overcomes reliance on predicting padding tokens by adding limited variability in the number of possible padding tokens, with some success; however, results still pale in comparison to a model with perfect sequence length prediction.
    \item Additional techniques to sample categorical diffusion models must be developed in order to improve sequence diversity and coverage. Compared to the traditional applications of diffusion models in image generation, molecular SMILES generation has significantly fewer possible outputs, making the models more likely to converge to a single prediction. This effect is heightened by the the presence of multiple SMILES strings mapping to the same molecular structure: while the sequence itself may be different, the same molecule is produced. We must develop methods to encourage greater output diversity in the sampled molecules. \ours attempts encourages output diversity using an ensemble approach, and while the results are significantly more diverse than single model approaches, the number of differently sampled molecules remains low.
\end{enumerate}
The case studies of \ours additionally demonstrate directions of future research for retrosynthetic modeling as a whole, many of which are echoed in other work \cite{chen2023g}:
\begin{enumerate}
    \setcounter{enumi}{2}
    \item Machine learning models for retrosynthesis are generally unaware of possible side products as a result of the generally available training data. Because these models are most commonly trained on patent reaction datasets, models are not generally exposed to explicit examples where multiple possible outcomes could occur. Indeed, the UPSTO-50K dataset has no reactant sets which map to more than one molecule, limiting the ability of the model to learn that multiple reactions may be possible for a single set of reactants. To improve the ability of ML models to understand chemical processes, incorporating additional examples into the training datasets may be beneficial, and with proper learning techniques, may lead to improved performance.
    \item Evaluating the performance of ML models on chemical retrosynthesis most commonly relies on a) matching to patented reactions and b) analyzing individual case studies. Evaluating models by matching to patented reactions limits the discovery of novel techniques and pathways, as models may produce valid results that do not match patent data. In contrast, evaluating individual case studies is more adaptable to understanding the model's true performance, but is a time consuming process requiring considerable chemical knowledge. To enhance the development of ML models for chemical retrosynthesis, we must develop new methods of analysis that take into account the abundance of viable reactions as well as existing chemical knowledge.
\end{enumerate}
Further research in these four tasks would greatly benefit the application of diffusion models in chemical retrosynthesis, as well as both diffusion modeling and retrosynthetic prediction models individually.

\subsection{Conclusion}

Our ensemble of categorical diffusion models provides state-of-the-art top-1 accuracy as well as competitive top-3, top-5, and top-10 accuracy for template-free methods. We demonstrate that our ensemble model is capable of not only reproducing patented reactions, but also producing viable alternative reactions for the same product. Additionally, we offer insight into the importance of length prediction and length variation methods when training diffusion models for sequence prediction tasks. Finally, we include an analysis and discussion of the limitations of our approach. In doing so, we hope to provide a suitable baseline and open new avenues of exploration for a new class of template-free retrosynthesis model which differs from current autoregressive approaches. In future work, we plan to investigate additional methods of length prediction, as well as apply different diffusion sampling techniques to improve the diversity of sampled molecules. Finally, we plan to continue improving upon the diffusion model that comprise \ours, introducing methods such as adaptive noise scheduling \cite{yuan2022seqdiffuseq} as well as other advancements in categorical diffusion models to improve model performance.





\section*{Declarations}

\bmhead{Availability of Data and Materials}

Our data and code is publicly available at \url{https://github.com/sfcurre/DiffER}.



\bmhead{Funding}

This project was made possible, in part, by support from the National Science Foundation grant no. IIS-2133650 and the National Library of Medicine grant no. 1R01LM014385. Any opinions, findings and conclusions or recommendations expressed in this paper are those of the authors and do not necessarily reflect the views of the funding agency.

\bmhead{Authors' Contributions}

SC designed, conceptualized, and implemented the methodology and wrote the manuscript. ZC assisted in reviewing the methodology and code. DAA assisted in the case study. All authors helped write the manuscript. All authors read and approved the final manuscript.














\begin{appendices}

\section{Overview of Retrosynthesis Models Used for Comparison}
\label{sec:A1}

We compare and report results against a variety of retrosynthesis models, as seen in Table \ref{tab:topk}. Models are grouped according to their model type, which includes template-based, semi-template, and template-free models. 

The template-based models we compare against are:
\begin{itemize}
    \item Retrosim \cite{coley2017computer}, which applies reaction templates derived from molecules similar to the target product, and ranks reactions by similarity to existing reactions; 
    \item Neuralsym \cite{segler2017neural}, which utilizes multi-layer perceptrons with molecular fingerprints of the product to predict applicable reaction templates;
    \item GLN \cite{dai2019retrosynthesis}, which uses graph neural networks to learn when and where in molecules reaction rules can be applied, while scoring the feasibility of the reaction;
    \item LocalRetro \cite{chen2021deep}, which focuses on the local environment of atoms in the product and classifies reaction templates on an atomic level.
\end{itemize}
Each of these methods utilize reaction templates sourced from known reactions. The primary goals are to 1.) select where in the molecule the reaction will take place and 2.) select which reaction template is applicable.

The semi-template models we compare against are:
\begin{itemize}
    \item G2Gs \cite{shi2020graph}, which predicts reactions centers and sequentially completes synthons through the use of a variational graph autoencoder;
    \item GraphRetro \cite{somnath2021learning}, which uses message-passing neural networks to classify reaction centers and completes synthons by predicting which leaving groups to add from among a pre-selected set;
    \item RetroXpert \cite{yan2020retroxpert}, which uses edge-enhanced graph attention networks to predict reaction centers and completes synthons using a transformer network;
    \item RetroPrime \cite{wang2021retroprime}, which uses two separate transformers, one to predict reaction centers, and the other to map synthons to reactants;
    \item G$^2$Retro \cite{chen2023g}, which uses a message passing network to predict three different types of reaction centers and sequentially adds substructures on to the synthons until complete reactants are formed.
    \item GDiffRetro \cite{sun2025gdiffretro}, which uses a dual graph molecular representation to train a reaction center prediction model and uses a conditional diffusion model to complete the produced synthons. 
\end{itemize}
Each of these methods first divide the product into synthons and then transform the synthons into complete reactants. The primary goals are to 1.) select where in the molecule the reaction will take place and 2.) complete synthons formed from the divided product into complete molecules which will react as needed.

The template-free models we compare against are:
\begin{itemize}
    \item Seq2Seq \cite{liu2017retrosynthetic}, which applies sequence-to-sequnece encoder-decoder recurrent neural networks to predict reactant SMILES from the product SMILES;
    \item Levenshtein \cite{sumner2020levenshtein}, which augments the training datasets of sequence-to-sequence recurrent neural networks by ensuring that the source and target SMILES strings have similar subsequences;
    \item GTA \cite{seo2021gta}, which incorporates graphical information in a sequence-to-sequence model by limiting the self-attention layer using the adjacency matrix of the product molecular graph; 
    \item Graph2SMILES \cite{tu2022permutation}, which leverages the permutation invariant nature of graph structures to remove variance that occurs when product is represented in SMILES form, and leverages transformer architectures to predict the reactant SMILES;
    \item Dual-TF \cite{sun2021towards}, which unifies graph-based and sequence-based methods to learn an energy-based model which ranks possible reactants according to their energy score;
    \item MEGAN \cite{sacha2021molecule}, which models one-step retrosynthesis as a series of graph edits, and trains a encoder-decoder graph attention model to how to adapt the product in a set of reactants;
    \item Chemformer \cite{irwin2022chemformer}, which pretrains a transformer architecture on SMILES strings using a masking approach and fine-tunes the model on product and reactant strings for retrosynthesis;
    \item Retroformer \cite{wan2022retroformer}, which jointly processes the molecular sequence and graph and uses localized attention to relay information between the reaction center and global chemical context when constructing reactants;
    \item Tied transformer \cite{kim2021valid}, which uses two-way transformers to encourage diversity and grammatical accuracy in predicted SMILES strings;
    \item R-SMILES \cite{zhong2022root}, which restructures SMILES representations of products and reactants to have significant structural overlap, and then uses neural machine translation architectures to map from products to reactants.
\end{itemize}
Most of these methods utilize transformer architectures on SMILES strings, effectively translating from a product string into a reactant string. Many methods also directly incorporate information from the molecular graph, augmenting the SMILES strings with known structural information. The primary goal is to directly predict a representation of the reactant from the product, while secondary goals are to ensure the validity and viability of the predicted reactant representation.





\end{appendices}


\bibliography{sn-bibliography}

\end{document}